\documentclass{article} 
\usepackage[submission]{colm2025_conference}
\colmfinaltrue
\usepackage{microtype}
\usepackage{hyperref}
\usepackage{url}
\usepackage{booktabs}
\usepackage{float}
\usepackage{lineno}
\usepackage{graphicx}
\usepackage{subcaption}
\usepackage{array}
\usepackage{amsmath}
\usepackage{multicol}

\definecolor{darkblue}{rgb}{0, 0, 0.5}
\hypersetup{colorlinks=true, citecolor=darkblue, linkcolor=darkblue, urlcolor=darkblue}

\title{TextBandit: Evaluating Probabilistic Reasoning in LLMs Through Language-Only Decision Tasks}

\author{Jimin Lim$^{\star}$ \hspace{2 cm} Arjun Damerla \hspace{2 cm} Arthur Jiang\\ UC Merced \hspace{2 cm} UC Berkeley \hspace{2.45 cm} Algoverse\\ \href{mailto:jlim85@ucmerced.edu}{jlim85@ucmerced.edu} \hspace{0.4cm} \href{mailto:arjundamerla@berkeley.edu}{arjundamerla@berkeley.edu} \hspace{0.10cm} \href{mailto:arthur@chainedtears.dev}{arthur@chainedtears.dev} \\ \\ \textbf{Nam Le} \hspace{2.5cm} \\ Algoverse \\ \href{mailto:nam.le94568@gmail.com}{nam.le94568@gmail.com} 
}%

\begin{document}

\ifcolmsubmission
\fi

\maketitle

\begin{abstract}
Large language models (LLMs) have shown to be increasingly capable of performing reasoning tasks, but their ability to make sequential decisions under uncertainty only using natural language remains underexplored. We introduce a novel benchmark in which LLMs interact with multi-armed bandit environments using purely textual feedback, “you earned a token”, without access to numerical cues or explicit probabilities, resulting in the model to infer latent reward structures purely off linguistic cues and to adapt accordingly. We evaluated the  performance of four open-source LLMs and compare their performance to standard decision-making algorithms such as Thompson Sampling, Epsilon Greedy, Upper Confidence Bound (UCB), and random choice. While most of the LLMs underperformed compared to the baselines, Qwen3-4B, achieved the best-arm selection rate of 89.2\% , which significantly outperformed both the larger LLMs and traditional methods. Our findings suggest that probabilistic reasoning is able to emerge from language alone, and we present this benchmark as a step towards evaluating decision-making capabilities in naturalistic, non-numeric contexts.
\end{abstract}

\section{Introduction}
Large Language Models (LLMs) have shown some Bayesian-like reasoning in simple tasks, it is still unknown if they are able to handle complex uncertainty with just natural language description. This ability could allow for more flexible and accessible approaches to decision making under uncertainty. Decision-making under uncertainty is used throughout many areas, but traditional methods such as Bayesian inferences and reinforcement learning often require complex math and data that may not be readily available. Recent studies show that LLMs exhibit Bayesian-like behavior in constrained tasks \cite{gupta2025coinflips}; \cite{felicioni2024uncertainty}, but it remains unclear on if they are able to generalize this ability to multi-step decision contexts which involve adapting through results and reasoning under uncertainty. Despite recent advances in LLMs such as GPT-4 and Llama-3.1-8B which have demonstrated strong language-based reasoning and zero-shot tasks, their ability to handle complex uncertainty relying on only natural language remains unclear. To address this, we introduce our method TextBandit, which is a novel benchmark that is designed to evaluate whether large language models are able to make sequential decision under uncertainty using only natural language feedback. To our knowledge, there is no prior benchmark that evaluates LLMs in this manner. Our setup, runs a suite of natural language bandit simulation tasks that measure the model’s ability to learn and make decisions based solely on text-based feedback, giving responds like "you earned a token". Four transformer-based open-source models are evaluated across 500 trials of bandit games with reward structures that vary, which measures their adaption and decision-making over time. We then compare the results from the LLMs against standard probabilistic baselines such as Epsilon-greedy, UCB, and Thompson sampling. After the LLM behavior is compared against the baselines, it shows that Qwen3-4B achieved the best-arm selection rate of 89.2\% which significantly outperformed all classical methods. We find that current LLMs are decent at making decisions under uncertainty when facing natural language descriptions as they can achieve similar scores as human methods. Larger models tend to take longer than other models and their results still fall short of smaller ones due to overthinking. In natural language bandit simulations, the results suggest that when a model thinks longer, it leads to mediocre or worse decision-making.

\section{Related Work}

\textbf{Probabilistic Reasoning in LLMs} This research is constructed of multiple key points in the study of LLMs, connecting their abilities to reason to theories of decision-making and evaluation frameworks. The research into whether LLMs can perform probabilistic reasoning is a central theme. Recent works have shown and explored the extent to which LLMs can mimic formal probabilistic models, most commonly Bayesian Inference. \cite{xie2022explanation} frames in-context learning as a form of implicit Bayesian inference, characterizes how LLMs can carry out posterior prediction by inferring and averaging latent concepts, although there are differences in the prompts, and pretraining data.\cite{gupta2025coinflips} demonstrate that while LLMs may have inherent priors, they can update their beliefs to be consistent with Bayesian posterior updates when provided with enough in-context evidence - suggesting how LLM's abilities for probabilistic reasoning surpasses simple pattern matching. \cite{sun2025llmenhanced} proposed that with integrated of classic bandit strategies and LLM-based reward prediction, it resulted in improved performance over direct LLM arm-selection in setting that had minimal semantic cues, which supports the design rationale behind our TextBandit approach.

\textbf{Uncertainty-Aware Decision-Making}
\cite{felicioni2024uncertainty} explores the benefits of the explicit consideration of epistemic uncertainty in the performance of LLMs in sequential decision-making. Demonstrating that LLMs can explore and adapt better in uncertainty-aware environments, the study infuses uncertainty-aware strategies, like posterior sampling, into the model. This makes the point that uncertainty is not merely a constraint, but can be exploited as a valuable signal to direct more effective and flexible model behavior - especially in probabilistic environments such as those considered in our benchmark (bandit environments).

\textbf{Exploration-Exploitation in Bandit Environments} Previous studies have examined the exploration-exploitation (E\&E) strategies of LLMs that are used in simulations under uncertainty. 
\cite{zhang2025llmhumanbandits}  compares the strategies used by LLMs to human methods such as the Upper Confidence Bound (UCB) algorithm to uncover the LLM's ability to simulate human behavior using the context of multi armed bandit simulations. Their findings reveal the impact of reasoning on exploration, the differences in E\&E behaviors between human methods and LLMs, as well as interpretations on how LLMs can be utilized for dynamic decision-making tasks. Specifically, the LLMs tested have been exploring more options in the beginning than at the end of the evaluation. Human methods explored more with diverse tactics such as random or direct methods and managed to achieve low regret. When Chain-Of-Thought is applied to these models, the reasoning capability increases dramatically, where they behave similarly to human methods. 

\section{Benchmark Design}
We propose a novel benchmark that evaluates LLMs in decision-making tasks under uncertainty using a multi-armed bandit (MAB) framework. The bandit environment consists up of multiple arms, each with a reward distribution that is unknown, and the goal is for the agent to identify the arm that maximizes cumulative reward over time. Unlike traditional setups that utilize numeric feedback, our benchmark requires the LLMs to infer latent reward structures purely off textual feedback. More specifically, the LLMs are provided with feedback after each decision \textit{“you earned a token”} for choosing the correct option and \textit{“you did not earn a token”} for the unsuccessful one. The challenge is that the models are not provided with explicit probabilities or numerical cues, hence requiring them to adapt based on linguistic cues alone. This experiment uses two arms that are fixed but have unknown success rates, one having a 30\% success rate while the other having 65\% success rate. Over a series of multiple iterations, the models must select one arm per round and adjust based on the feedback they receive. The performance of each model is evaluated based on the cumulative reward, regret, and best-arm selection rate. In addition to the LLMs, we will evaluate several decision-making algorithms typically seen in multi-armed bandit problems, including Epsilon greedy, Upper Confidence Bound (UCB), Thompson Sampling, and Random Choice. These algorithms will serve as baselines that allow us to compare the performance of the LLMs against well established decision-making strategies. Each of these algorithms will help in comparing the LLMs’ ability to adapt to feedback and in maximizing the cumulative reward and minimizing cumulative regret over time.

\section{Methodology}
\subsection{Task Overview and Reward Structure}
In our benchmark, we simulate a multi-armed bandit environment where large language models (LLMs) must make repeated decisions under uncertainty using only natural language feedback. Each bandit environment consists of multiple arms (ranging from 2 to 5), with each arm associated with a fixed but unknown success probability. For example, in the 2-arm configuration, one arm yields a reward with a 65\% probability and the other with 30\%. These probabilities are never revealed to the model.

At each round, the LLM is prompted to select an arm. Based on the sampled outcome, the model receives textual feedback:
\begin{itemize}
  \item "You earned a token" if the action results in success (reward = 1)
  \item "You did not earn a token" if it results in failure (reward = 0)
\end{itemize}
No explicit numerical cues or probabilistic information are provided. Importantly, there are no penalties for incorrect choices: The only signal the model receives is whether it succeeded or failed, in linguistic form. The objective of the model is to maximize cumulative reward across multiple iterations by learning which arm is better solely from this binary language feedback.

\subsection{Prompting Protocol}
Each LLM is evaluated using a consistent prompting structure designed to simulate a text-only decision-making loop. The core prompt consists of:
\begin{itemize}
  \item A natural language instruction that puts the task into the context of decision-making situation (e.g. Such as, “Select the slot machine that you think will yield you a token.”)
  \item A history of previous choices and their outcomes in plain language (e.g., “Slot machine 1 won,” “Slot machine 2 lost”), spanning all prior iterations in the current episode
  \item A request for the model to select the next action by outputting a number corresponding to the arm (e.g., “1”, “2”, “3”, etc.)
\end{itemize}\
The model receives this prompt anew at each step, with the historical context updated to reflect the outcomes of previous choices. No internal memory of past interactions is preserved between runs. Each decision is made in a single-shot completion with no intermediate reasoning or Chain-of-Thought scaffolding. To ensure consistency, we apply the same format and structure across all models and arm configurations. The only variation lies in the number of arms available and the accumulated outcome history. This protocol isolates the model’s ability to infer and adapt to reward patterns based solely on linguistic reinforcement, rather than numeric data or structured training signals.

\subsection{Baselines and Comparison Models}
In order to test the LLM's ability with this dataset, we compared it with many models that are commonly used in bandit decision-making research. Random Choice chooses actions at random, without learning. \cite{do2024epsilon} selects the best possible action using the probability $1 - \epsilon$, otherwise it will choose at random  \cite{russo2020thompson} uses Bayesian inference to collect information about the probability distribution within the bandit simulation, sampling from those distributions to make decisions. \cite{hao2019bootstrapucb} analyzes the options provided and will utilize the more successful options while continuing to try new ones.

\subsection{LLMs Evaluated}
To test our hypothesis, we selected a diverse set of open-source large language models. The models we chose represent different architectures, parameter sizes, and different training methodologies, allowing for an extensive analysis of how these factors can influence an LLM’s decision-making abilities. The models evaluated in our benchmarks include Qwen3-4B, Qwen3-8B, Llama-3.1-8B, and Phi-2 .

Our experimental design uses the multi-armed bandit problem within a purely text-based interaction loop. For each trial, the LLM receives a prompt containing the history of its previous choices and outcomes (e.g., “Slot machine 1 won,” “Slot machine 2 lost”). The prompt explicitly instructs the model to act as a decision making agent and to only output the number (ID) of its chosen machine, “1” or “2” for example. This setup does not give the LLM information on the underlying reward structure - a 30\% win rate for slot machine 1 and a 65\% win rate for slot machine 2. Evaluation is conducted over 500 independent runs, with each run consisting of 25 decision making iterations. This repetitive process allows us to access the model’s ability to learn and adapt its strategy over time. Performance is measured via best-arm selection rate, which will track its frequency of being chosen over the objectively inferior machine (Slot machine 1). 

\begin{table}[H]
\centering
\caption{Large language models evaluated on the bandit task, along with architecture and key characteristics.}
\label{tab:llm-evaluated}
\resizebox{\textwidth}{!}{
\begin{tabular}{p{2.5cm} p{1.5cm} p{3cm} p{6.5cm}}
\toprule
\textbf{Model} & \textbf{Parameters} & \textbf{Architecture} & \textbf{Notable Characteristics}\\
\midrule
Qwen3-4B & 4B & decoder-only transformer & supports multilingual input, strong performance in reasoning tasks \\
Qwen3-8B & 8B & decoder-only transformer & larger version of Qwen3-4B, enhanced tool-use abilities, better for long-context understanding\\
Llama-3.1-8B & 8B & decoder-only transformer & optimized for following instructions and multilingual capabilities \\
Phi-2 & 2.7B & transformer & strong performance for its size, compact and efficient \\
\bottomrule
\end{tabular}
}
\end{table}

\section {Results}
Our evaluation of the LLMs on natural language-based multi-armed bandit tasks revealed significant differences in performance and results across the different tested architectures. We have found that models such as Qwen3-4B demonstrated their ability to learn and adapt over strategies to maximize rewards, while other models struggled to find the optimal arm.
    \subsection{Quantitative Performance}
    We assessed models based on three key metrics: Cumulative Reward, Best-Arm Selection Rate, and Cumulative Regret. These metrics provide insights on each model's decision-making and learning capability over 500 independent runs of 25 iterations each. To calculate cumulative reward, we add a token for receiving a successful outcome and not adding anything when receiving the failed outcome. 

    \subsection{Cumulative Reward}
\begin{figure}[H]
    \centering
    \begin{subfigure}[b]{0.48\linewidth}
        \includegraphics[width=\linewidth]{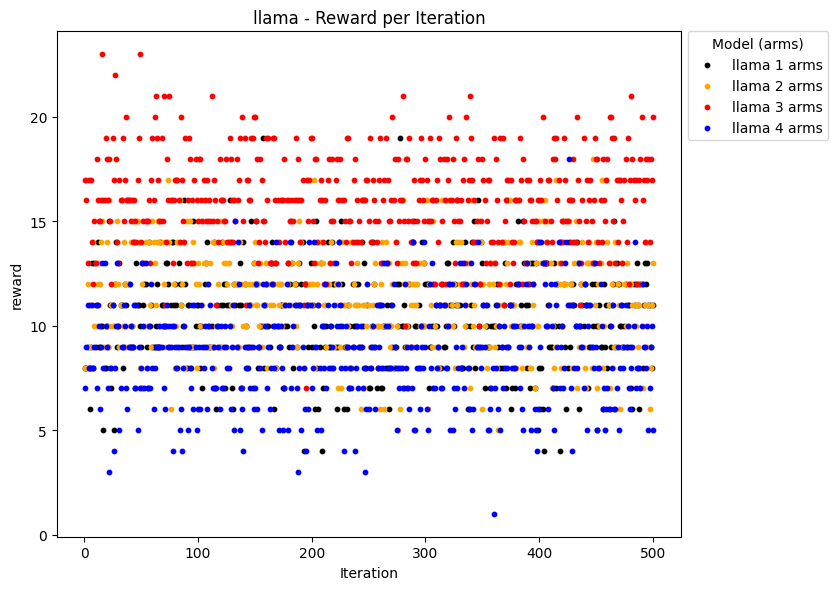}
        \caption{Llama-3.1-8B reward trends: Shows scattered performance and inconsistent preference across trials}
        \label{fig:llama-reward}
    \end{subfigure}
    \hfill
    \begin{subfigure}[b]{0.48\linewidth}
        \includegraphics[width=\linewidth]{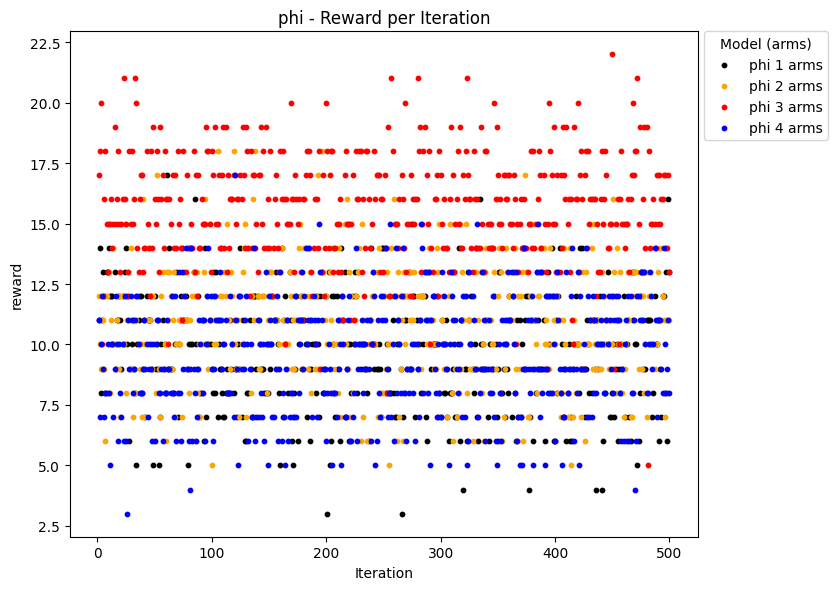}
        \caption{Phi-2 reward trends: Displays highly random behavior with poor learning over time.}
        \label{fig:phi-reward}
    \end{subfigure}
    
    \vspace{0.5cm} 
    
    \begin{subfigure}[b]{0.48\linewidth}
        \includegraphics[width=\linewidth]{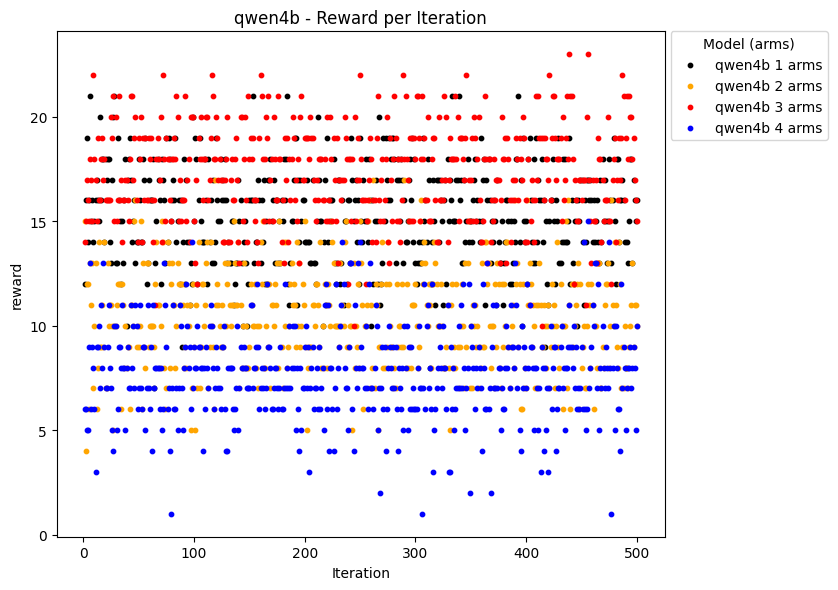}
        \caption{Qwen3-4B reward trends: Demonstrates consistent preference for the optimal arm with high cumulative reward.}
        \label{fig:qwen4b-reward}
    \end{subfigure}
    \hfill
    \begin{subfigure}[b]{0.48\linewidth}
        \includegraphics[width=\linewidth]{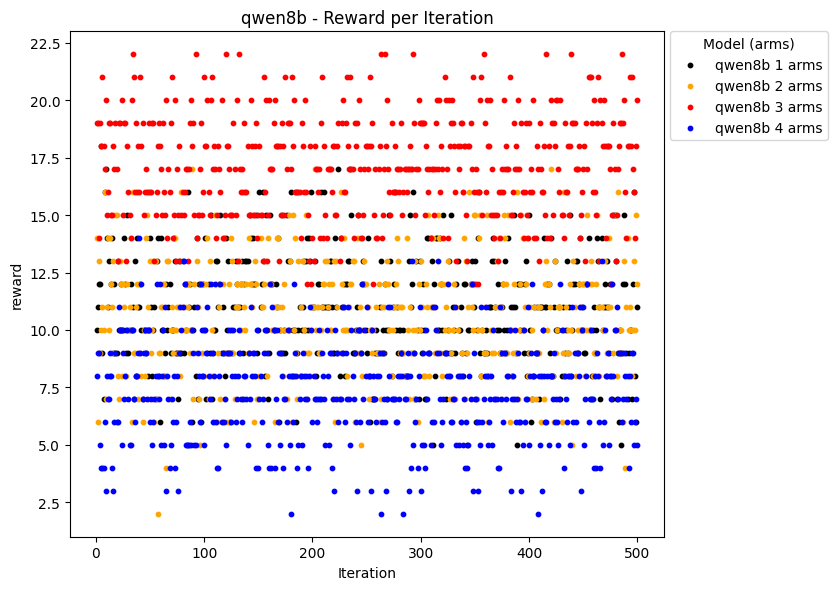}
        \caption{Qwen3-8B reward trends:Signs of overthinking and instability, resulting in suboptimal decisions.}
        \label{fig:qwen8b-reward}
    \end{subfigure}
    
    \caption{Comparison of reward trends across four LLMs in the bandit task.}
    \label{fig:reward-comparison}
\end{figure}

        The cumulative reward illustrates the total number of tokens the model accumulated over the 25 decision-making iterations.  Surprisingly, the \hyperref[fig:qwen4b-reward]{Qwen3-4B} model shows more accuracy when choosing the optimal arm, therefore accumulating the most amount of tokens with the highest rewards rate. In contrast, \hyperref[fig:llama-reward]{Llama-3.1-8B}, \hyperref[fig:phi-reward]{Phi-2}, and \hyperref[fig:qwen8b-reward]{Qwen3-8B's} amount of total reward accumulated is substantially lesser, suggesting it's performance closer to random chance and a failure to consistently choose the better arm. 
    \subsection{Cumulative Regret}
\begin{figure}[H]
    \centering
    \begin{subfigure}[b]{0.48\linewidth}
        \includegraphics[width=\linewidth]{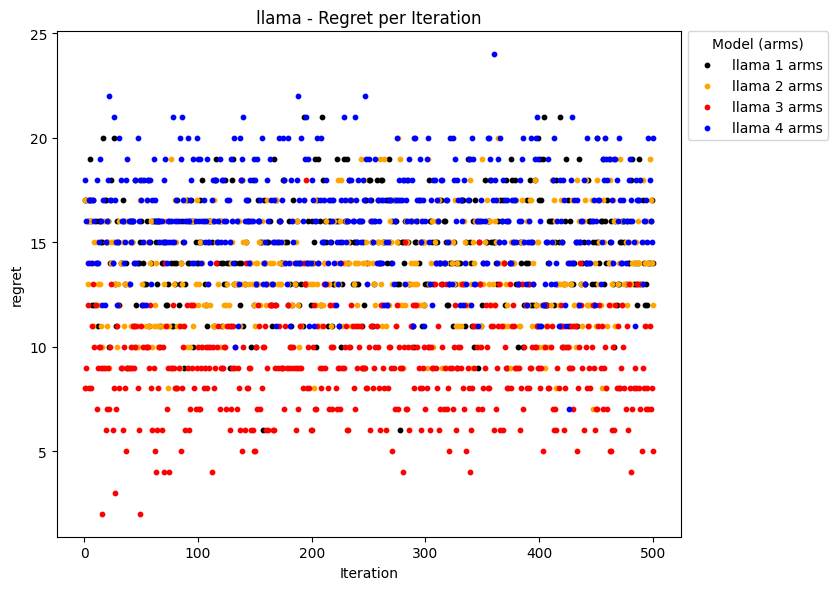}
        \caption{Llama-3.1-8B regret trends: Exhibits high cumulative regret, suggesting poor adaptation to feedback over time.}
        \label{fig:llama-regret}
    \end{subfigure}
    \hfill
    \begin{subfigure}[b]{0.48\linewidth}
        \includegraphics[width=\linewidth]{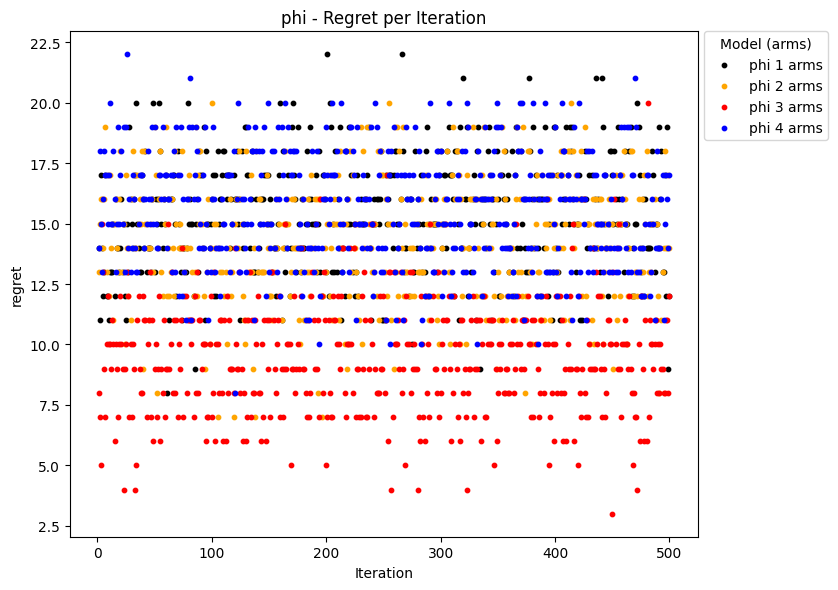}
        \caption{Phi-2 regret trends: Maintains consistently high regret levels, indicating limited learning from outcomes}
        \label{fig:phi2-regret}
    \end{subfigure}
    
    \vspace{0.5cm}
    
    \begin{subfigure}[b]{0.48\linewidth}
        \includegraphics[width=\linewidth]{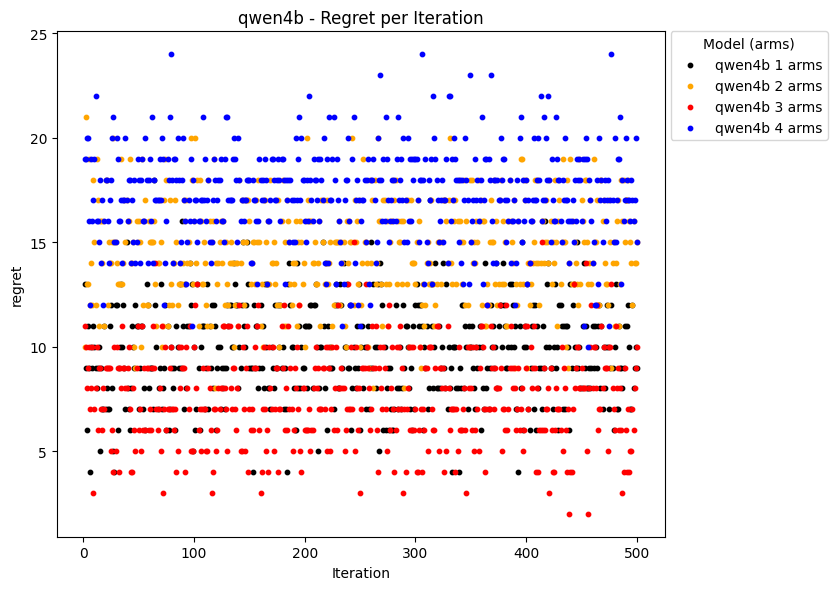}
        \caption{Qwen3-4B regret trends: Displays rapid reduction in regret, reflecting strong and consistent decision making}
        \label{fig:qwen4b-regret}
    \end{subfigure}
    \hfill
    \begin{subfigure}[b]{0.48\linewidth}
        \includegraphics[width=\linewidth]{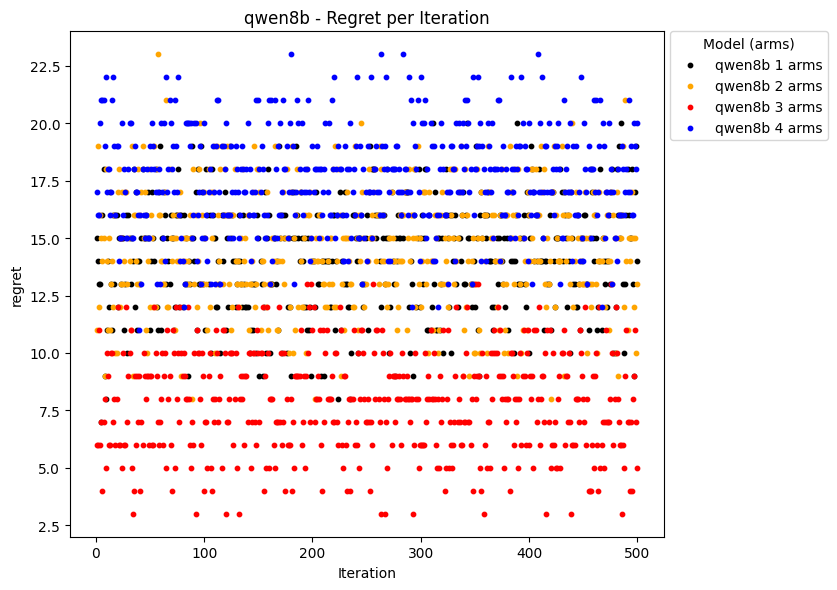} 
        \caption{Qwen3-8B regret trends : Consistently high regret across prompts, indicating overthinking and difficulty in identifying optimal arm, despite larger model size}
        \label{fig:qwen8b-regret}
    \end{subfigure}

    \caption{Comparison of cumulative regret trends for four LLMs.}
    \label{fig:regret-comparison}
\end{figure}

        Cumulative regret, shown in \autoref{fig:regret-comparison}, measures the opportunity cost of not choosing the optimal arm. A lower cumulative regret signifies a more efficient decision making process. The regret trends is very similar to what's shown in Cumulative Reward. An unexpected turnout is that the prompt with four arms, had the lowest amounts of cumulative regret across all models while the prompt with five arms, had the highest amounts of cumulative regret. \hyperref[fig:llama-regret]{Llama-3.1-8B} and \hyperref[fig:phi2-regret]{Phi-2}'s regret scores are varied across the same prompts, indicating that it has a low capacity for probabilistic reasoning when under uncertainty. Qwen3-4B has similar patterns to the rest for the prompts with three and five arms, but excel when there are two arms. This suggests that due to it's smaller size, it thinks faster and manages to exploit the optimal arm.

    \subsection{Best-Arm Selection Rate}
        The best arm selection rate, shown in table, quantifies the percentage of times each model chose the arm with the 65\% success rate (the optimal arm). Qwen3-8B, Llama-3.1-8B, and Phi-2 models achieved best-arm selection rates of 37.5\%, 31.6\%, and 25.4\%, respectively. These rates are considerably the lower and indicate a struggle to distinguish the better-performing arm from the inferior ones. Despite Qwen3-8B's tendency to overthink, it still manages to achieve better results than the other models meaning that some of its decisions are still valid. Phi-2 is also a smaller model similar to Qwen3-4B, but it achieved the worst results out of all the models.

\subsection{Comparison to Baselines}
We compared the performance of LLMs with four standard multi-armed bandit baselines, Thompson Sampling, Upper Confidence Bound (UCB), Epsilon-Greedy and Random Choice in order to have all these performances in the same context. These baselines can utilize structured decision-making heuristics restricted to probabilistic decision-making, but are not able to make use of language comprehension. On the other hand, the LLMs work entirely on natural language feedback, and the comparison was a result of emergent probabilistic reasoning on language itself. 

The most appropriate metrics to evaluate learning rates was the performance in best-arm selection rate, which measures the percentage of the time an agent picked the best arm. Among the baselines, Thompson Sampling had the best best-arm selection rate of 51.1\%, UCB the second-best of 47.6\%, Epsilon-Greedy the third-best of 38.1\% and Random Choice in last with 31.8\%.

However, a single LLM, Qwen3-4B, performed much better than all baselines in terms of a best-arm selection rate that equaled 89.2\%, which signifies that the LLM was advanced in the capabilities of learning linguistic feedback and developing a consistent policy of maximizing rewards. The other LLMs Qwen3-8B (37.5\%), Llama-3.1-8B (31.6\%) and Phi-2 (25.4\%) did worse than both Thompson Sampling and UCB, indicating their inability to appropriately adapt to the task, or to make sense out of the feedback.

\begin{table}[H]
\begin{center}
\begin{tabular}{lcccc}

\toprule
\multicolumn{1}{c}{\bf MODEL}
&\multicolumn{1}{c}{\bf FINAL CUMULATIVE REWARD }
&\multicolumn{1}{c}{\bf BEST-ARM SELECTION RATE}\\
\midrule
Thompson-Sampling & 8297 & 51.1\% \\
UCB               & 4696 & 47.6\% \\
Epsilon-Greedy    & 6029 & 38.1\% \\
Random-Choice     & 5783 & 31.8\% \\
\bottomrule
\end{tabular}
\end{center}
\caption{Performance of various Baselines on natural language multi-armed bandit tasks.}\label{model-performance}
\end{table}

\subsection{Qualitative Analysis}
The LLMs that are tested usually use Chain-Of-Thought, but in these datasets it is removed to receive a clear output. As a result, they follow similar patterns like when a random option is chosen at first, they will try to exploit that option despite it not having the best win rate. 
Their method is similar to the Thompson Sampling method, where they balance exploration with exploitation. The LLMs will sample the options first and choose the ones they believe are the most successful. This leads them to choosing some optimal options but not the most optimal one because they believe that their chosen one is the best after receiving a moderate amount of outputs. 
Notably, Qwen3-8B took an exceptionally large amount of time when testing because it kept trying to reason instead of giving a concise input. 
\begin{table}[H]
\begin{center}
\begin{tabular}{lcccc}

\toprule
\multicolumn{1}{c}{\bf MODEL}
&\multicolumn{1}{c}{\bf FINAL CUMULATIVE REWARD }
&\multicolumn{1}{c}{\bf BEST-ARM SELECTION RATE}\\
\midrule
Qwen3-4B         & 11150 & 89.2\% \\
Qwen3-8B         & 4686 & 37.5\% \\
Llama-3.1-8B     & 3946 & 31.6\% \\
Phi-2            & 3181 & 25.4\% \\
\bottomrule
\end{tabular}
\end{center}
\caption{Performance of various LLMs on natural language multi-armed bandit tasks.}\label{model-performance}
\end{table}

\section {Discussion}
In comparison to the baselines, the LLMs reasoning capabilities are inferior with the exception Qwen3-4B. This suggests they may have developed different biases on which choices they believed had the best probabilities while the baselines, methods that don't involve reason, were able to reach decisions on this dataset because they were optimized for better probabilistic calculation and problem-solving. Our findings show that some LLMs, most notably Qwen3-4B, have the flexibility to adapt to uncertainty using natural language alone, with significant differences between models. This suggests that purely off-language-based interactions, basic probabilistic reasoning form without the use of numerical cues. Models such as Qwen3-8B and Llama-3.1-8B, which were larger, struggled to consistently identify the optimal arm. This suggests that there is no correlation between model size and making better decisions in this context. In fact, the base Qwen3-4B and Qwen3-8B models received identical pretraining and has similar qualities besides the amount of parameters so the training is not a cause for this difference either. It may be that the architecture of Qwen-4B, as an efficient and lightweight model, contributed to its impressive probabilistic reasoning in this fast-feedback environment, where they receive limited information. Larger models may have been trained for complex reasoning which is why when they encounter simpler tasks, they tend to overthink things which leads to a drop in performance. Although models such as Qwen3-8B and Llama-3.1-8B have a greater capacity for abstract reasoning, their under performance may be resulted from overfitting to irrelevant features as shown in the feedback prompt of excessive internal deliberation. Similar patterns have been seen in \cite{zhang2025llmhumanbandits} where the LLMs halted their exploration as they received more information, which leads them to solidify an optimal arm from noise. Their training done on complex reasoning may introduce biases in simple reinforcement environments like ours. Compared to medium sized models like Qwen3-4B which appear to utilize a more direct exploitation strategy, resulting in better performance. The idea that smaller models having higher performance in terms of internal probabilistic reasoning is unlikely as the smallest model tested, Phi-2, produced the worst outcome. Another possibility is that the LLMs have an internal bias where they make an answer based on the input given without any deep reasoning such as choosing a specific option based on the example prompts they received. This answer may be different from their internal reasoning, so they over-complicate their thoughts and deviate their decisions from their calculations. With the amount of resources the larger LLMs have, they suffer more heavily from this behavior and generate repetitive content, preventing them from providing a final answer. Unlike in \cite{zhang2025llmhumanbandits}, the LLMs were restricted from using Chain-Of-Thought which suggests why their performance unremarkable. Without Chain-Of-Thought, their pure probabilistic reasoning ability is low. This behavior is more pronounced in larger models, Qwen3-8B is an example of this as despite the vast amount of time it spent thinking, it's performance was only mediocre. While some models could learn effectively from text-based feedback, the others behaved in a much more random manner and lacked a robust internal strategy. Some examples of further work include the implementation of more complex tasks, such as dynamic tasks or multi-step reasoning to further evaluate and develop the probabilistic capabilities of LLMs.

\section {Limitations and Future Direction}
Although our findings showed that certain LLMs seem to exhibit strong probabilistic reasoning using purely natural language, there are several limitations to be noted. First, our benchmark was limited to a small set of open-source models, which are primally between 2.7B and 8B parameters. Larger-scale models such as Qwen-32B or GPT-4 were not included due to computation limits. It is possible that these larger models may exhibit different behaviors, such as a more stable converge or superior adaption under uncertainty. Second, due to the benchmark's reward structure being static and simple it does not take into account of non-stationary environments, delayed feedback, or multi-step planning that is seen in real-world decision-making tasks. Future work could expand upon having more complex environments to test the horizontal reasoning as well as dynamic adaption. Third, our prompting strategy avoided chain-of-thought in order to isolate the single-shot decision capabilities. This allowed us to study the raw language-based adaptation but may have led to underestimating what the LLMs are capable of with more guided prompting. Future directions include testing in non-stationary or dynamic bandits environments with delayed rewards, incorporating chain-of-thought prompting in order to stuff how scaffolds affect the exploration-exploration balance, or using larger models with fine-tuned variants to evaluate scaling trends.

\section {Conclusion}
We introduced TextBandit, a benchmark in evaluating the abilities of large language models in making decisions in uncertain environment with only the guidance of natural language alone. By framing the multi-armed bandit problem with a natural language task, we have found that LLMs have a decent capacity for successful judgment when under uncertainty and influenced by natural language. Our evaluations show that the LLM's size does not translate to better performance. In fact, it may return results that are less effective. TextBandit offers a minimal yet challenging benchmark that shows another perspective in the evaluation of and adaptation of language modes. With this benchmark, we can contribute to deeper understandings of probabilistic reasoning for LLMs under uncertainty as well as information that can be used to create opportunities for the further development of this ability. 

\section {Ethics Statement}
Our study did not involve human subjects, private data, or any interventions in living individuals; all experiments conducted were performed on synthetic bandit tasks with publicly available open source LLMs. 

\section {Reproducibility Statement}
We release all the code, evaluation scripts, and open-source models that were used in our experiments at \url{https://github.com/ChainedTears/TextBandit}. The repository contains detailed documentation on the models that were used, the environment setup instructions, and how to reproduce the results. All experiments rely on open-source LLMs available with the Hugging Face Transformers library, and were conducted using GPU instances hosted on RunPod, which allowed for reproducibility without access to local high-end hardware.

\appendix
\section{Appendix}
    \subsection{Prompting}
    All tests were conducted with a "fair" or "equal" system prompt; consisting of an equal number of multi-shot prompting per example, as follows:

    \begin{minipage}[b]{0.45\textwidth}
        \centering
        \includegraphics[width=\textwidth]{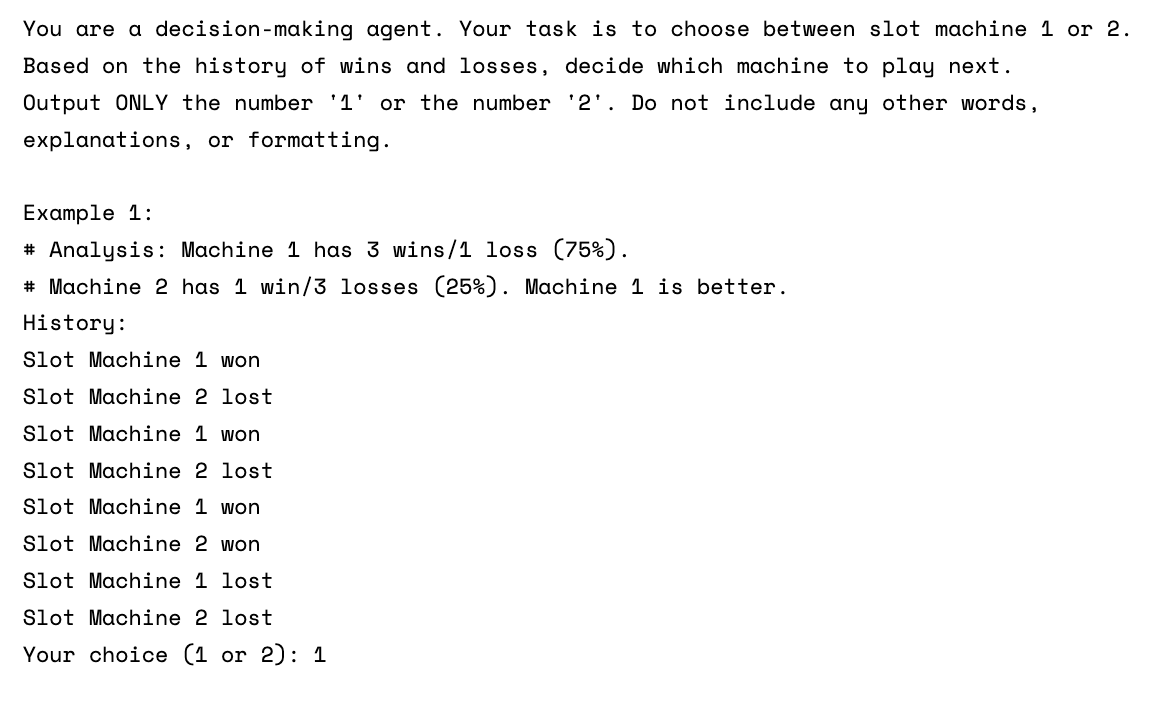}
        \label{fig:image1}
    \end{minipage}
    \hfill
    \begin{minipage}[b]{0.45\textwidth}
        \centering
        \includegraphics[width=\textwidth]{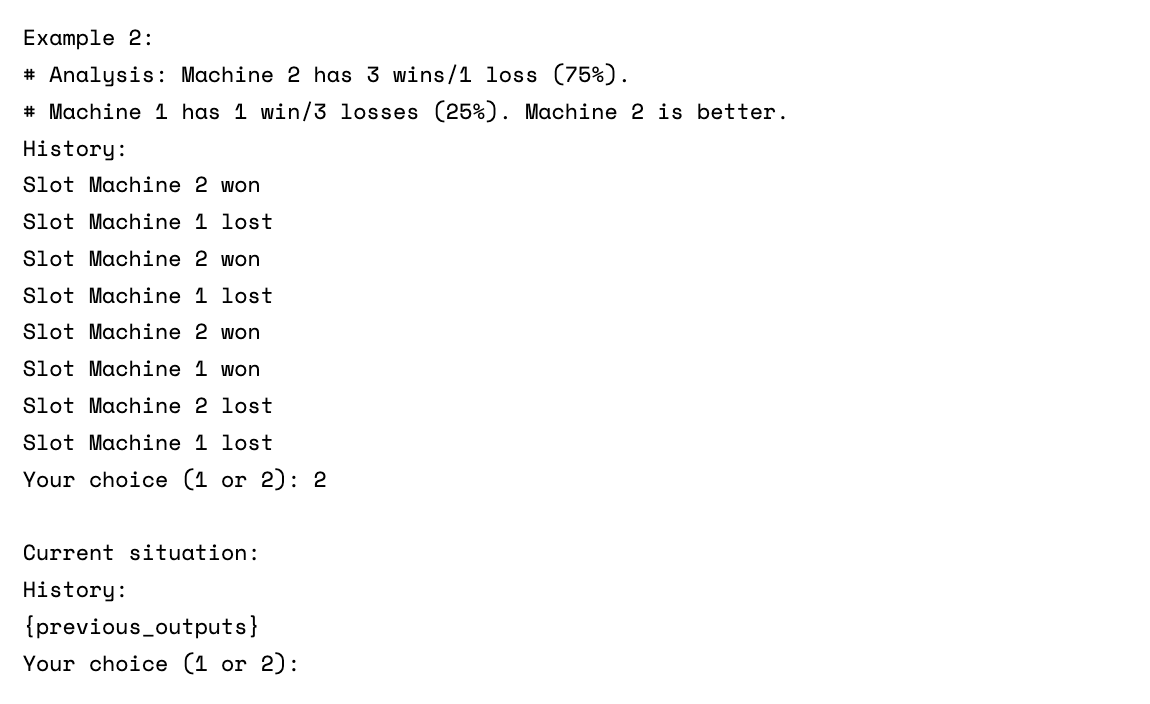}
        \label{fig:image2}
    \end{minipage}

    The amount of examples grows along with the quantity of arms present.
    \subsection{Implementation Details}
    This implementation consists of setting up an LLM as a decision-making agent using various pre-trained open-source language models. The core concept revolves around asking an AI to select the optimal slot machine based on a history of wins and losses. For each round, the AI is presented with the past outcomes of playing different slot machines. It then generates a response, from which its chosen machine (a single digit) is extracted. The chosen machine's outcome (win or loss) is simulated based on a predefined probability which is not shown to the AI, and this new outcome is added to the history for the next decision. The system evaluates the AI's performance by tracking how often it selects the objectively "best" slot machine (the one with the highest win rate) and its cumulative reward over multiple iterations.
    \subsection{Important Data Points}
    Here is some information crucial to understanding our Research Paper:
    \begin{itemize}
    \item It has become evident that more arms would indicate a decrease in the model's performance and accuracy when predicting the optimal arm. As data show us, Llama-3.1-8B's average accuracy of choosing the optimal arm went from 31.56\% during the two arm test to 7.37\% on the five arm test. This trend is observed throughout all models; with Phi-2 going from 25.45\% to  17.78\%, Qwen3-4B from 89.22\% to 6.53\%, and Qwen3-8B from 37.49\% to 17.09\%.
    \item Another phenomenon we observed was how the performance and accuracy of the models were worsened progressively as the models grew larger, starting from 4 billion parameters. Evidence reveals how Qwen3-4B, the second smallest model with four billion parameters, was the best performing LLM out of all the models we have tested, with an average accuracy in all tests of 44.56\%, with other models gradually worsening: Qwen3-8B (8 billion parameters) at 33.96\%, Llama-3.1-8B (8 billion parameters) at 30.7\%, and Phi-2 (2.7 billion parameters) at 27.6\%. This establishes how Phi-2 was too small to be good at predicting the optimal arm, while Llama-3.1-8B and Qwen3-8B were too big and overthinking too much to provide a practical answer.
    \end{itemize}

\bibliography{colm2025_conference}

\begin{thebibliography}{8}
\providecommand{\natexlab}[1]{#1}
\providecommand{\url}[1]{\texttt{#1}}
\expandafter\ifx\csname urlstyle\endcsname\relax
  \providecommand{\doi}[1]{doi: #1}\else
  \providecommand{\doi}{doi: \begingroup \urlstyle{rm}\Url}\fi

\bibitem[Do et~al.(2024)Do, Adebiyi, and Zhang]{do2024epsilon}
Bach Do, Taiwo Adebiyi, and Ruda Zhang.
\newblock Epsilon-greedy thompson sampling to bayesian optimization.
\newblock \url{https://arxiv.org/pdf/2403.00540}, October 2024.
\newblock URL \url{https://arxiv.org/pdf/2403.00540}.
\newblock University of Houston.

\bibitem[Felicioni et~al.(2024)Felicioni, Maystre, Ghiassian, and Ciosek]{felicioni2024uncertainty}
Nicolò Felicioni, Lucas Maystre, Sina Ghiassian, and Kamil Ciosek.
\newblock On the importance of uncertainty in decision-making with large language model.
\newblock \url{https://arxiv.org/html/2404.02649}, April 2024.
\newblock URL \url{https://arxiv.org/html/2404.02649}.
\newblock Politecnico di Milano and Spotify. Licensed under CC BY 4.0.

\bibitem[Gupta et~al.(2025)Gupta, Corona, Ge, Wang, Klein, Darrell, and Chan]{gupta2025coinflips}
Ritwik Gupta, Rodolfo Corona, Jiaxin Ge, Eric Wang, Dan Klein, Trevor Darrell, and David~M. Chan.
\newblock Enough coin flips can make llms act bayesian.
\newblock \url{https://arxiv.org/pdf/2503.04722}, March 2025.
\newblock URL \url{https://arxiv.org/pdf/2503.04722}.
\newblock University of California, Berkeley.

\bibitem[Hao et~al.(2019)Hao, Abbasi-Yadkori, Wen, and Cheng]{hao2019bootstrapucb}
Botao Hao, Yasin Abbasi-Yadkori, Zheng Wen, and Guang Cheng.
\newblock Bootstrapping upper confidence bound.
\newblock \url{https://arxiv.org/pdf/1906.05247}, October 2019.
\newblock URL \url{https://arxiv.org/pdf/1906.05247}.
\newblock Purdue University, VinAI, DeepMind.

\bibitem[Russo et~al.(2020)Russo, Van~Roy, Kazerouni, Osband, and Wen]{russo2020thompson}
Daniel~J. Russo, Benjamin Van~Roy, Abbas Kazerouni, Ian Osband, and Zheng Wen.
\newblock A tutorial on thompson sampling.
\newblock \url{https://arxiv.org/pdf/1707.02038}, July 2020.
\newblock URL \url{https://arxiv.org/pdf/1707.02038}.
\newblock Columbia University, Stanford University, Google DeepMind, Adobe Research.

\bibitem[Sun et~al.(2025)Sun, Wang, Yang, Xiao, Lui, and Dai]{sun2025llmenhanced}
Jiahang Sun, Zhiyong Wang, Runhan Yang, Chenjun Xiao, John C.~S. Lui, and Zhongxiang Dai.
\newblock Large language model–enhanced multi‑armed bandits, February 2025.
\newblock URL \url{https://arxiv.org/pdf/2502.01118}.

\bibitem[Xie et~al.(2022)Xie, Raghunathan, Liang, and Ma]{xie2022explanation}
Sang~Michael Xie, Aditi Raghunathan, Percy Liang, and Tengyu Ma.
\newblock An explanation of in-context learning as implicit bayesian inference.
\newblock \url{https://arxiv.org/pdf/2111.02080}, July 2022.
\newblock URL \url{https://arxiv.org/pdf/2111.02080}.
\newblock Stanford University.

\bibitem[Zhang et~al.(2025)Zhang, Wang, Chen, Mansur, and Sarhangian]{zhang2025llmhumanbandits}
Ziyuan Zhang, Darcy Wang, Ningyuan Chen, Rodrigo Mansur, and Vahid Sarhangian.
\newblock Comparing exploration–exploitation strategies of llms and humans: Insights from standard multi-armed bandit tasks.
\newblock \url{https://arxiv.org/pdf/2505.09901}, May 2025.
\newblock URL \url{https://arxiv.org/pdf/2505.09901}.
\newblock University of Toronto.

\end{thebibliography}
\bibliographystyle{colm2025_conference}

\end{document}